\documentclass[runningheads]{llncs}

\usepackage[T1]{fontenc}
\usepackage{graphicx}
\usepackage{cite}
\usepackage{booktabs}
\usepackage{multirow}
\usepackage{amsmath}
\usepackage{verbatim}

\begin{document}

\title{Breaking Down the Hierarchy: A New Approach to Leukemia Classification} 

\author{Ibraheem Hamdi\inst{1}\ \and
Hosam El-Gendy\inst{1}\and
Ahmed Sharshar\inst{1} \and
Mohamed Saeed\inst{1} \and
Muhammad Ridzuan\inst{1} \and 
Shahrukh K. Hashmi\inst{2} \and
Naveed Syed\inst{2} \and
Imran Mirza\inst{2} \and
Shakir Hussain\inst{2} \and
Amira Mahmoud Abdalla\inst{2} \and
Mohammad Yaqub\inst{1}}

\authorrunning{I. Hamdi et al.}

\institute{Mohamed Bin Zayed University of Artificial Intelligence, Abu Dhabi, UAE \\ \email{\{ibraheem.hamdi, hosam.elgendy, ahmed.sharshar, mohamed.saeed, muhammad.ridzuan, mohammad.yaqub\}@mbzuai.ac.ae} \and Sheikh Shakhbout Medical City, Abu Dhabi, UAE \\ \email{\{shhashmi, nasyed, immirza, shahussain, amirabdalla\}@ssmc.ae}
}

\maketitle             

\begin{abstract}
The complexities inherent to leukemia, multifaceted cancer affecting white blood cells, pose considerable diagnostic and treatment challenges, primarily due to reliance on laborious morphological analyses and expert judgment that are susceptible to errors. Addressing these challenges, this study presents a refined, comprehensive strategy leveraging advanced deep-learning techniques for the classification of leukemia subtypes. We commence by developing a hierarchical label taxonomy, paving the way for differentiating between various subtypes of leukemia. The research further introduces a novel hierarchical approach inspired by clinical procedures capable of accurately classifying diverse types of leukemia alongside reactive and healthy cells. An integral part of this study involves a meticulous examination of the performance of Convolutional Neural Networks (CNNs) and Vision Transformers (ViTs) as classifiers. The proposed method exhibits an impressive success rate, achieving approximately 90\% accuracy across all leukemia subtypes, as substantiated by our experimental results. A visual representation of the experimental findings is provided to enhance the model's explainability and aid in understanding the classification process.
\end{abstract}


\section{Introduction}

Leukemia is a type of cancer that starts in the bone marrow and spreads to the blood. Patients commonly exhibit symptoms such as fatigue, weakness, frequent infections and more \cite{mayo_clinic_leukemia}. Leukemia ranks among the ten most common cancers, with hundreds of thousands of new cases and deaths every year \cite{huang2022disease}.

There are two main leukemia categories based on the speed of progression: acute, which develops quickly and exhibits symptoms within weeks of forming, and chronic, which develops more slowly and may not show noticeable symptoms for years. Based on the type of blood cell affected, these can be further divided into two subcategories: myeloid and lymphocytic~\cite{mayo_clinic_leukemia}. Therefore, leukemia can generally be divided into four main types: acute lymphocytic leukemia (ALL), acute myeloid leukemia (AML), chronic lymphocytic leukemia (CLL), and chronic myeloid leukemia (CML).

The diagnosis of leukemia typically involves physical examinations, medical history evaluations, and laboratory tests. Blood sample analysis and bone marrow biopsies are common methods for leukemia diagnosis. Bone marrow biopsies are invasive and uncomfortable for patients \cite{biopsy}. In contrast, blood smears are non-invasive but provide less information and require skilled pathologists, who are in short supply \cite{bychkov_schubert_2023}. Moreover, diagnosing leukemia from slides is a manual process, subject to bias and operator errors \cite{mohapatra2010image}.

This paper investigates the potential of using computer vision algorithms to classify leukemia and its subtypes from blood smear images and proposes a novel hierarchical solution. This task is largely unexplored in the literature at the moment, therefore, an effective solution is needed to enhance accuracy, reduce costs of further testing, and reduces the use of invasive methods. The principal contributions of this work are as follows:

\begin{itemize}
    \item Development of a hierarchical deep learning method to accurately classify different types of leukemia.    
    \item To our knowledge, we are the first to investigate the classification of a wide range of leukemia types alongside reactive and healthy cells.
    \item In-depth analysis of the performance of CNNs and ViTs on the classification of leukemia.
\end{itemize}


\section{Related Work}
Previous work focused on detecting and segmenting leukemia cells from microscopic blood smear images \cite{dhal2020acute,genovese2021acute,das2021transfer,mohapatra2012lymphocyte}, while some worked on classifying a specific type of leukemia against healthy cells~\cite{jothi2019rough, shah2019classification, negm2018decision, rawat2017computer}. However, at the time of writing, the authors are only aware of two papers that work on classifying multiple types against healthy cells, and both target the ALL, AML, CLL, and CML subtypes.

The authors in~\cite{ahmed2019identification} propose using a small custom CNN, using the ALL-IDB~\cite{6115881} dataset as well as some images from the American Society of Hematology (ASH) Image Bank~\cite{imagebank}. The authors obtained 903 images and implemented several augmentations to reduce overfitting. Despite this, their results show a large difference between the training and validation accuracies, indicating that their model is greatly overfitting. This can be attributed to the small dataset and overly simplified model. In the most recent paper~\cite{9425264}, the authors propose using a pre-trained GoogleNet model, which is fine-tuned on 1,200 images from the ASH Image Bank. The authors do not provide detailed information on the hyperparameters used, making it difficult to study their impact on the model.

In summary, both papers lacked sufficient detail regarding the techniques used to attain their results. 
CNNs utilized were either small or obsolete, having been surpassed in performance by newer models such as ConvNeXt \cite{liu2022convnet} or Vision Transformers (ViTs) \cite{dosovitskiy2020image}.


\section{Dataset}

The dataset consists of slices from 84 slides, each representing a patient. It was collected by a local hospital using $1000\times$ zoom on each slide. The slides were divided into patches which were annotated by a senior pathologist to obtain ground-truth labels.

The dataset, which includes four main leukemia types, normal cells, acute promyelocytic leukemia (APML), and reactive cells, is unique and represents the Middle East, making it the first of its kind for the region. APML is an aggressive, rare subtype of AML, requiring a specific treatment approach \cite{apl_treatment}. Reactive cells are non-cancerous cells that appear similar to leukemia cells \cite{hamad2019lymphocytosis}. 

\subsection{Pre-Processing}
Many steps were taken to refine and clean the data in preparation for training. Specifically, scans with less than four images were excluded, and duplicates were removed. Furthermore, most images had a resolution of $1200\times1600$; therefore, images with significantly lower resolution were excluded, producing 3710 images. These images have been resized to $384\times384$. Training and validation sets were created by splitting the data per slide to prevent any leakage and prevent the model from under-performing on unseen data.


\section{Methodology Overview}

\subsection{Flat/Leaf Classification (Baseline)}

\begin{figure}[!ht]
  \centering
  \includegraphics[width=\linewidth]{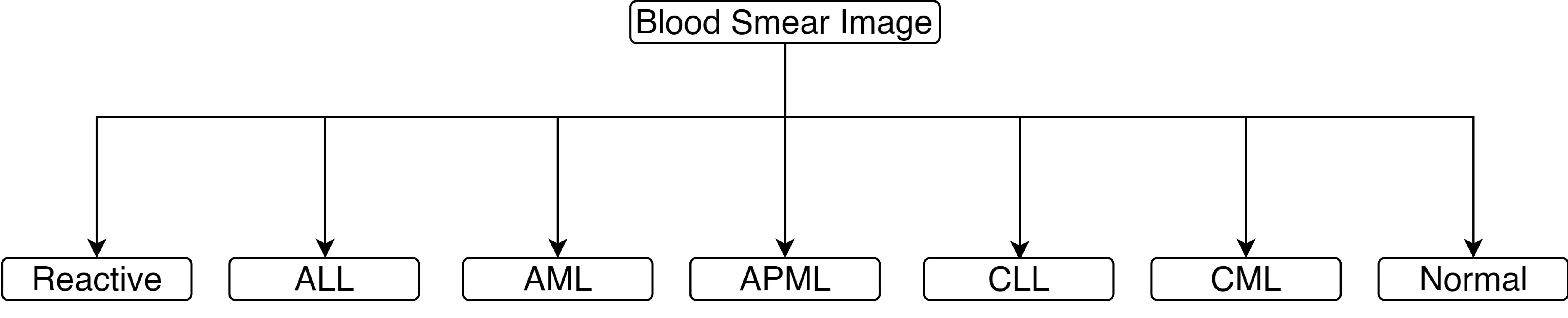}  
  \caption{Flat/Leaf Classification Structure.}
  \label{fig:leaf}
\end{figure}

This approach is referred to as Flat/Leaf Classification, as it focuses on all 7 "leaf" classes seen in Figure \ref{fig:leaf}. The approach is simple and ignores the hierarchy of the classes, which may lead to a loss of information and a potential reduction in performance. In addition, it does not provide classification probabilities for broader classes, such as acute vs chronic which is clinically vital in cases where doctors struggle to decide on the exact leukemia types.

\vspace{-15pt}

\subsection{Hierarchical Multi-Label Classification}
\begin{figure}[t]
  \centering
  \includegraphics[width=\linewidth]{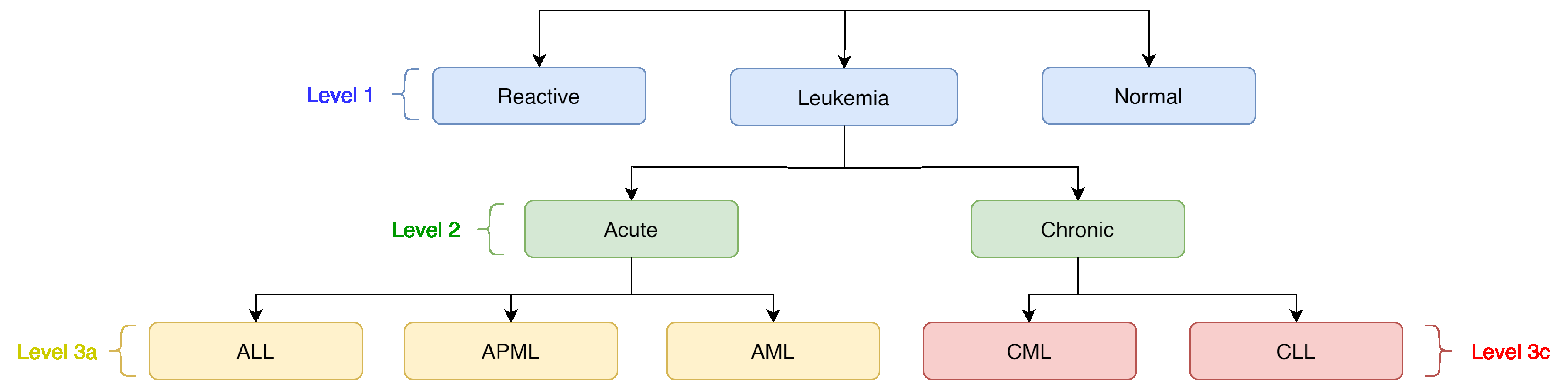}  
  \caption{Hierarchical structure of leukemia subtypes, showcasing different levels.}
  \label{fig:hierarchy}
\end{figure}

Another way of presenting the dataset is shown in Figure~\ref{fig:hierarchy}, providing a more intuitive structure in a tree-like diagram. This inspired the authors to develop a method that performs classification following this hierarchy. Since the chance of having multiple types of leukemia is unlikely~\cite{george2022mixed}, the authors assume non-overlapping subgroups in the data and produce a solution based on that. We proposed two different methods to apply the hierarchical architecture:
\begin{itemize}
    \item \textbf{Base Hierarchical Model:} Four separate models were trained; each on an individual level of the tree shown in Figure~\ref{fig:hierarchy}. They were then merged during inference to produce the final model. APML was included as a separate class in our study. This is because APML is unique among leukemia types in its excellent response to treatment, requiring immediate attention and intervention.
    \\
    \item \textbf{Proposed Multi-label Hierarchical  Model:} We've developed a hierarchical deep learning model for image classification, operating sequentially and tiered. The model first differentiates among normal, reactive, and leukemia cases(level 1). If leukemia is detected, it further classifies it as acute or chronic (level 2) and subsequently refines these into subtypes(level 3). These levels are shown in Figure~\ref{fig:hierarchy}. We employ a unique loss control strategy to ensure that the loss at each level is self-contained, preventing cross-layer interference. In essence, each stage of the hierarchy is affected only by its respective loss, promoting precise classifications throughout the model.
\end{itemize}


\section{Experimental Setup}

\subsection{Model Selection}

We utilize ConvNeXt-Tiny and ViT-Small models as they performed best, as shown in Table \ref{tab:baselines}. Note that these results validate our choice of these models, which were later tweaked to match the proposed model hyper-parameters.
 For the baseline, the chosen configurations included the Adam optimizer \cite{kingma2014adam}, Focal Loss \cite{lin2017focal}, and weighted-cross entropy with a learning rate of $10^{-6}$.

\begin{table}[!ht]
\centering
\caption{Performance of different baselines on multi-class classification task}
\label{tab:baselines}
\begin{tabular}{@{}llc@{}}
\toprule
\textbf{Model} &  & \textbf{Macro-Averaged F1} \\ \midrule
ResNet-18      &  & 62.18                      \\
ResNet-50      &  & 58.32                      \\
GoogleNet      &  & 69.3                       \\
DenseNet-121   &  & 70.5                       \\
\textbf{ConvNeXt-Tiny}  &  & \textbf{78.56}                      \\
\textbf{ViT-Small}      &  & \textbf{79.93}                      \\ \bottomrule
\end{tabular}
\end{table}

\subsection{Simulating Pathological Evaluation}

The holistic examination process taken by pathologists was taken into consideration. Hence, this study introduces a post-processing methodology that aggregates the mode of all predictions per slide (patient) at the end of the validation or testing phases. This approach mirrors the pathologists' diagnostic process and mitigates the impact of minor misclassifications by the models. Consequently, the models yield a singular dominant label per slide, eliminating the occurrence of edge cases that each slide has only one label.

\subsection{Experimental Procedures}

We further modified the baseline configurations. Using Focal Loss function instead of weighted-cross entropy enhanced the model's performance. The Adam optimizer was replaced with AdamW with a learning rate of $1\times10^{-6}$, and each experimental run was limited to 80 epochs for the sake of time, and this was enough for the model to saturate.

\subsection{Evaluation Metrics}

Experiments were evaluated using accuracy and F1 score metrics. For hierarchical classification, the hierarchical precision (hP), recall (hR), and F1-score (hF) were computed using Equation \ref{eq:3}.

\begin{align} \label{eq:3}
hP=\frac{\sum_i\left|\hat{P}_i \cap \widehat{T}_i\right|}{\sum_i\left|\hat{P}_i\right|}
\hspace{1cm}hR=\frac{\sum_i\left|\widehat{P}_i \cap \widehat{T}_i\right|}{\sum_i\left|\widehat{T}_i\right|}
\hspace{1cm}\mathrm{hF}=\frac{2 * \mathrm{hP} * \mathrm{hR}}{\mathrm{hP}+\mathrm{hR}}
\end{align}

The set $\hat{P}{i}$ contains the most specific class or classes that are predicted for a given test example $i$ and all of its ancestor classes. Likewise, $\widehat{T}{i}$ is the set that includes the true most specific class or classes for that test example $i$ and all of its ancestor classes. These sets are used to compute summations over all test examples, and it is important to note that the measures being used are extended versions of the commonly used precision metrics \cite{silla2011survey}.

\section{Results and Discussion}

\begin{table}[b]
\caption{Comparison of performance using CNN and ViT on multi-class.}
\label{tab:multiclass}
\setlength{\tabcolsep}{1pt} 
{\fontsize{8}{12}\selectfont 

\begin{tabular}{@{}lccccccccccccc@{}}
\toprule
                                  & \multicolumn{6}{c}{\textbf{All Classes}}                                                                                                                                &           & \multicolumn{6}{c}{\textbf{Without Reactive}}                                                                                                                       \\ \cmidrule(r){1-7} \cmidrule(l){9-14} 
\multicolumn{1}{c}{\textbf{}}     & \multicolumn{3}{c|}{\textbf{ConvNeXt}}                                                                             & \multicolumn{3}{c}{\textbf{ViT}}                   & \textbf{} & \multicolumn{3}{c|}{\textbf{ConvNeXt}}                                                                             & \multicolumn{3}{c}{\textbf{ViT}}               \\ \cmidrule(lr){2-7} \cmidrule(l){9-14} 
\multicolumn{1}{c}{\textbf{Fold}} & \textbf{ACC}                  & \textbf{F1}                   & \multicolumn{1}{c|}{\textbf{AUROC}}                & \textbf{ACC}    & \textbf{F1}     & \textbf{AUROC} &           & \textbf{ACC}                  & \textbf{F1}                   & \multicolumn{1}{c|}{\textbf{AUROC}}                & \textbf{ACC} & \textbf{F1}    & \textbf{AUROC} \\ \cmidrule(r){1-7} \cmidrule(l){9-14} 
1                                 & 61.58                         & 62.26                         & \multicolumn{1}{c|}{91.17}                         & 67.73           & 67.08           & 92.05          &           & 70.30                          & 71.62                         & \multicolumn{1}{c|}{92.60}                          & 80.78        & 74.15          & 93.21          \\
2                                 & 77.02                         & 72.31                         & \multicolumn{1}{c|}{92.18}                         & 74.12           & 75.37           & 94.11          &           & 77.07                         & 72.01                         & \multicolumn{1}{c|}{92.20}                          & 84.11        & 84.96          & 94.85          \\
3                                 & 81.76 & 81.82 & \multicolumn{1}{c|}{90.59} & 87.89           & 86.79           & 95.56          &           & 87.07 & 87.21 & \multicolumn{1}{c|}{98.61} & 90.82        & 90.23          & 97.69          \\
4                                 & 79.27                         & 79.93                         & \multicolumn{1}{c|}{96.34}                         & 77.45           & 75.21           & 97.08          &           & 82.34                         & 82.98                         & \multicolumn{1}{c|}{97.45}                         & 85.55        & 85.54          & 98.16          \\
5                                 & 79.04                         & 77.29                         & \multicolumn{1}{c|}{95.83}                         & 79.93           & 77.01           & 93.50           &           & 83.27                         & 80.19                         & \multicolumn{1}{c|}{92.83}                         & 83.74        & 81.82          & 94.35          \\
Avg                               & 75.73                        & 74.72                        & \multicolumn{1}{c|}{93.22}                        & \textbf{77.42} & \textbf{76.29} & \textbf{94.46} &           & 80.01                         & 78.80                        & \multicolumn{1}{c|}{94.73}                        & \textbf{85.00}  & \textbf{83.34} & \textbf{95.65} \\
Std                               & 8.09                          & 7.83                          & \multicolumn{1}{c|}{2.68}                          & 7.43            & 7.03            & 1.93           &           & 6.50                          & 6.85                          & \multicolumn{1}{c|}{3.04}                          & 3.69         & 5.95           & 2.16           \\ \bottomrule
\end{tabular}}
\end{table}

\subsection{Flat/Leaf Classification}
The classification outcomes of all seven leaf classes within a flat structure are concisely delineated in Table \ref{tab:multiclass}. A performance comparison revealed the ViT-Small model's superior efficacy and accelerated convergence pace compared to the ConvNext-Tiny model.

Analyzing confusion matrices revealed challenges in classifying reactive cells, known to pathologists for their ambiguous state between "Normal" and "Leukemia". Experiments excluding these cells improved performance.

Table~\ref{tab:multiclass} shows performance variations in 5-fold cross-validation due to dataset slides and labeling discrepancies. The lower standard deviation for ViTs, compared to CNNs, in six and seven-class classifications support adopting ViTs.

\subsection{Base vs Proposed Hierarchical Classification}

Table \ref{tab:binary} presents the comparative analysis of Base and Proposed hierarchical models in ConvNext and ViT architectures. Each level underwent an intermediate evaluation. While the Base model excels in intermediate binary classification due to dedicated tuning, the Proposed model marginally surpasses in seven-class classification due to its automated learning and scalability without the additional overhead of training multiple models separately.

Interestingly, ConvNeXt outperforms ViT in internal levels, likely due to smaller training data sizes as ViT needs much more data for training; therefore, smaller models fit better. Nonetheless, the highest performance of 90.97\% was achieved with the ViT architecture in the Proposed model for all seven classes.

\begin{table}[]
\centering
\caption{Accuracy of Hierarchical models at different stages }
\label{tab:binary}
\begin{tabular}{lllllllll}
\hline
                               &  & \textbf{ConvNeXt} &  & \textbf{ViT}   &  & \textbf{ConvNeXt} &  & \textbf{ViT} \\ \cline{3-5} \cline{7-9} 
Subtask                        &  & \multicolumn{3}{c}{\textbf{Base}}      &  & \multicolumn{3}{c}{\textbf{Proposed}}    \\ \cline{1-1} \cline{3-5} \cline{7-9} 

Normal vs Reactive vs Leukemia &  & \textbf{95.41}    &  & 94.23     &  & 93.58      &  & 92.04      \\

Acute vs Chronic     &  & \textbf{94.75}    &  & 93.21          &  & 91.16        &  & 92.80  \\

ALL vs AML   vs APML     &  & 81.65        &  & \textbf{83.47} &  & 79.96       &  & 82.59      \\

CLL vs CML       &  & \textbf{97.02}    &  & 96.74          &  & 95.08         &  & 94.47        \\

All 7 Classes          &  & {89.34}    &  & 90.01        &  & 90.48        &  & \textbf{90.97}      \\ \hline 
\end{tabular}
\end{table}

Moreover, although the ViT yields high flat/leaf multi-class classification results, the ConvNext-Tiny outperforms the ViT in most binary experiments. This can be due to the reduction of samples for the ViT. ViTs require large quantities of data to obtain reasonable representations~\cite{lee2021vision}.

\subsection{Flat vs Proposed Hierarchical Classification}

The comparison between the Flat and Proposed Hierarchical models can be shown in Table~\ref{tab:comparison}. These results show that our hierarchical structure is boosting the model performance in terms of precision and recall compared to the flat structure. This is more evident in the "visually confusing" classes, such as ALL and Reactive classes, where the flat structure model struggles to differentiate these classes from other classes. However, when placed in a hierarchical structure, it becomes easier for the model to differentiate between these classes.

\begin{table}[]
\centering

\caption{Hierarchical and flat evaluation metric comparisons on the ViT-Small model}
\label{tab:comparison}
\begin{tabular}{lllllclll}
\hline
           &  & \multicolumn{3}{c}{\textbf{Flat/Leaf}}                                                                         & \textbf{} & \multicolumn{3}{c}{\textbf{Hierarchical}}                                                                            \\ \cline{3-5} \cline{7-9} 
\textbf{Class}      &  & \multicolumn{1}{c}{\textbf{Precision}} & \multicolumn{1}{c}{\textbf{Recall}} & \multicolumn{1}{c}{\textbf{F1}} & \textbf{} & \multicolumn{1}{c}{\textbf{Precision}} & \multicolumn{1}{c}{\textbf{Recall}} & \multicolumn{1}{c}{\textbf{F1}} \\ \cline{1-1} \cline{3-5} \cline{7-9} 
ALL        &  & 0.66                                   & 0.42                                & 0.52                            &           & 0.98                                   & 0.88                                & \textbf{0.93}                   \\
AML + APML &  & 0.75                                   & 0.91                                & 0.82                            &           & 0.93                                   & 0.95                                & \textbf{0.94}                   \\
CLL        &  & 0.86                                   & 0.63                                & 0.73                            &           & 0.98                                   & 0.95                                & \textbf{0.97}                   \\
CML        &  & 0.93                                   & 0.93                                & 0.93                            &           & 0.97                                   & 0.96                                & \textbf{0.96}                   \\
Reactive   &  & 0.58                                   & 0.79                                & 0.67                            &           & 0.85                                   & 0.95                                & \textbf{0.89}                   \\ \hline
\end{tabular}
\end{table}

\subsection{Visual Experimental Results}
\begin{figure}[!ht]
\centering

\begin{tabular}{ccc}
 \includegraphics[width=0.29\linewidth,height=2.5cm]{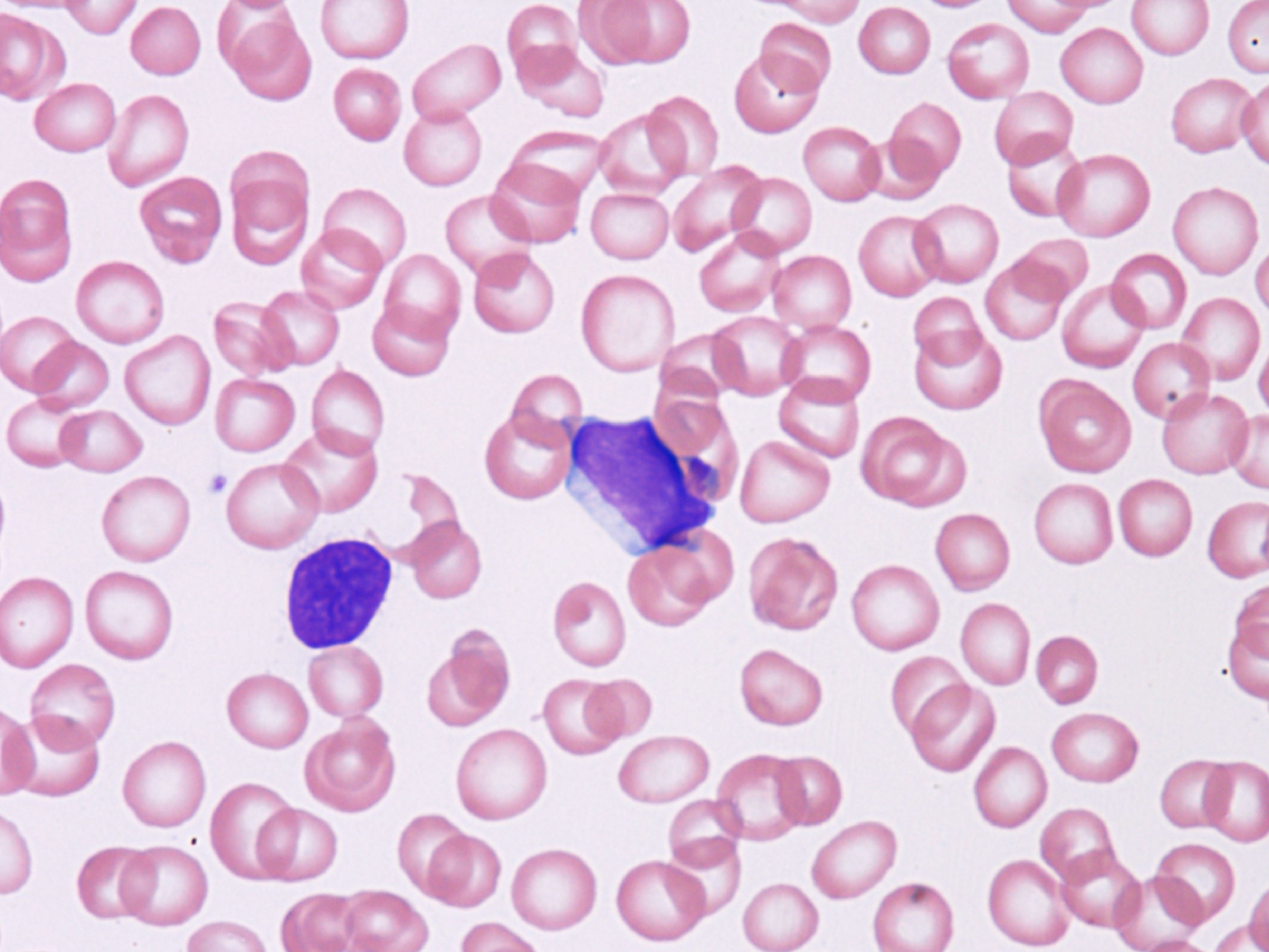}  &   
 \includegraphics[width=0.29\linewidth,height=2.5cm]{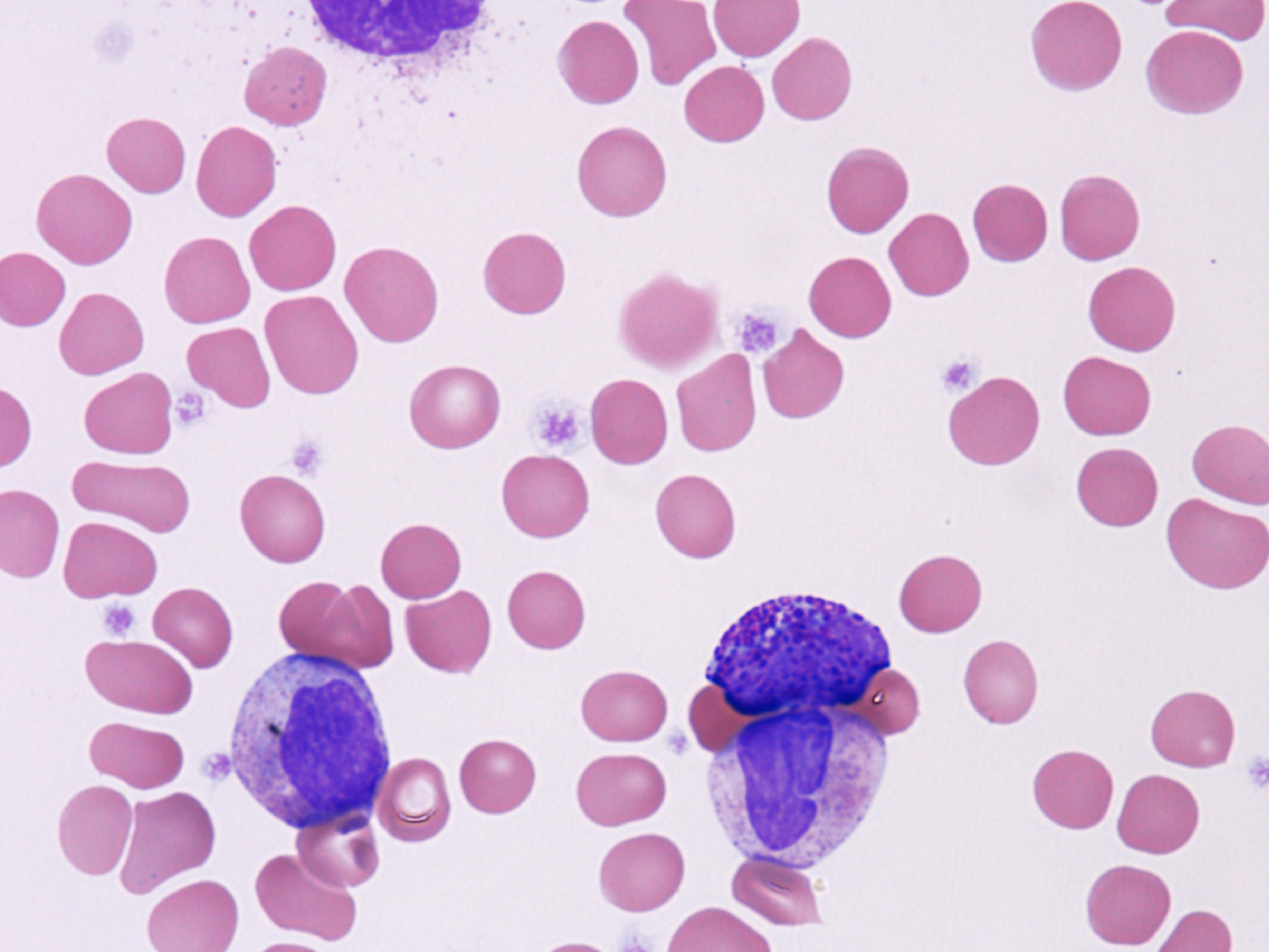}   &   
 \includegraphics[width=0.29\linewidth,height=2.5cm]{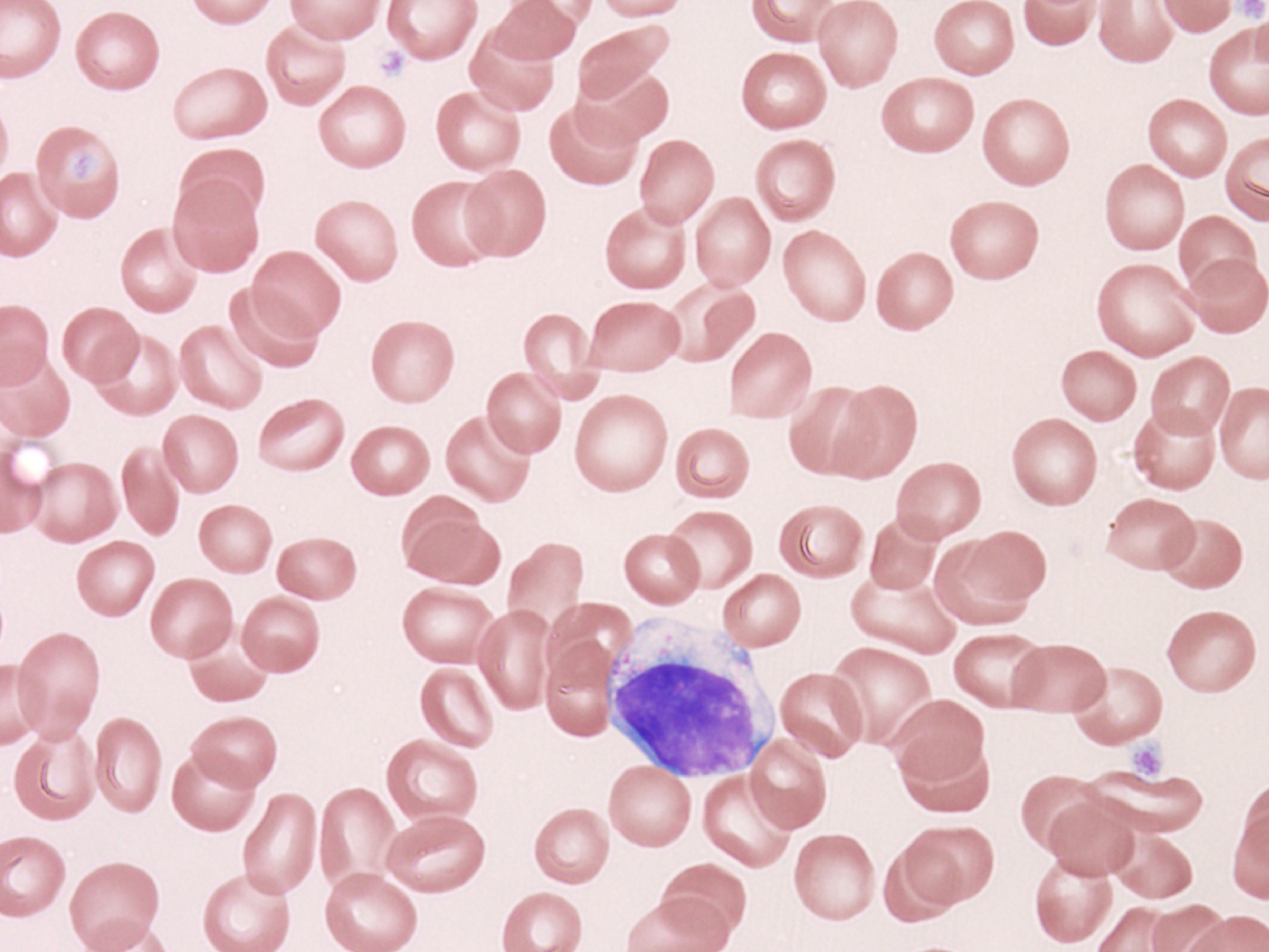} \\
\end{tabular}

\begin{tabular}{ccc}
 \includegraphics[width=0.29\linewidth,height=2.5cm]{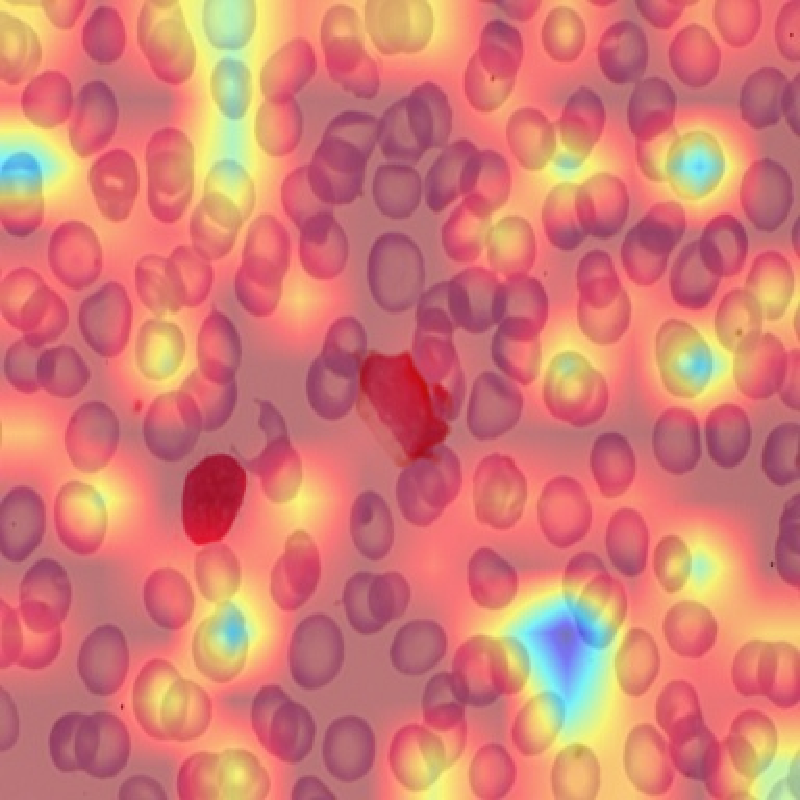}  &   
 \includegraphics[width=0.29\linewidth,height=2.5cm]{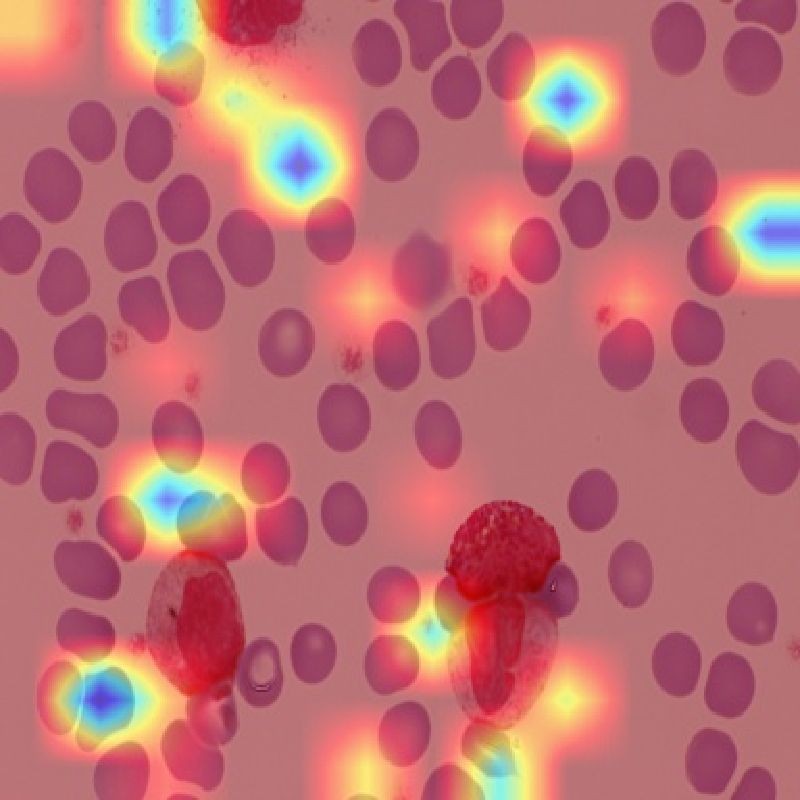}   &   
 \includegraphics[width=0.29\linewidth,height=2.5cm]{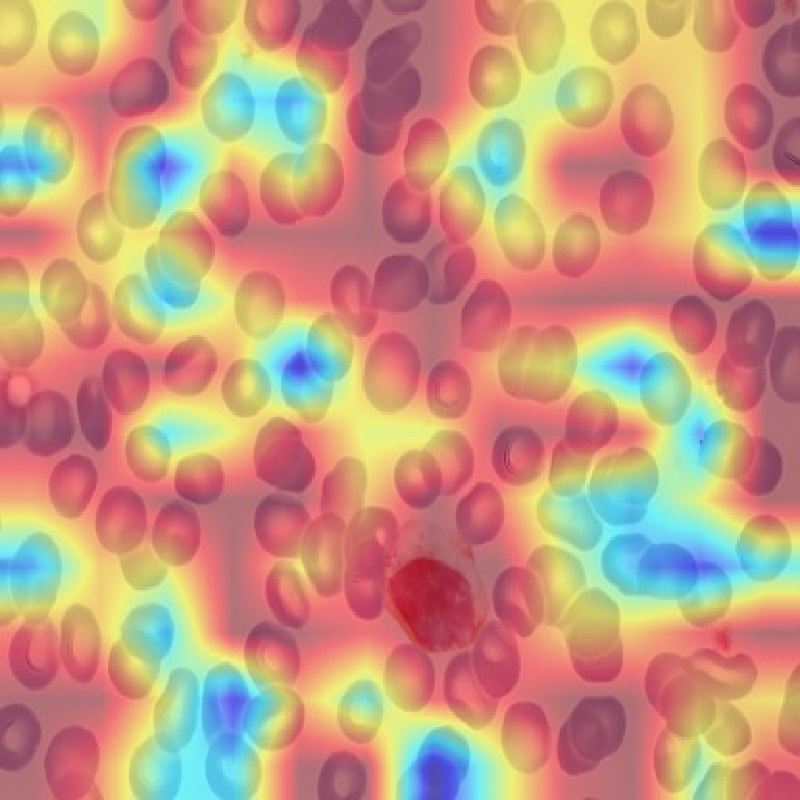} \\
\end{tabular}

\begin{tabular}{ccc}
 \includegraphics[width=0.29\linewidth,height=2.5cm]{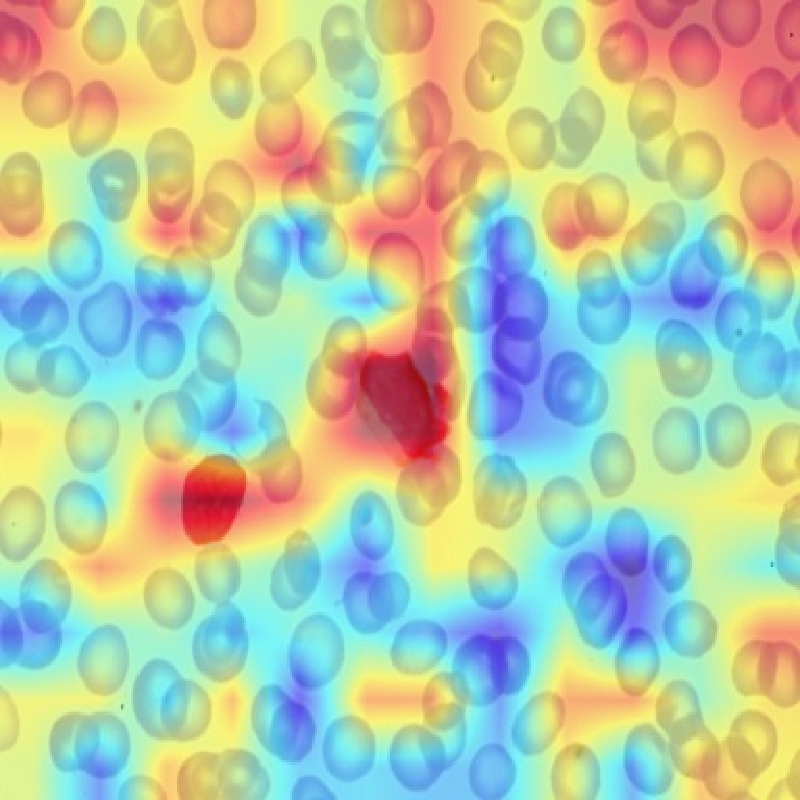}  &   
 \includegraphics[width=0.29\linewidth,height=2.5cm]{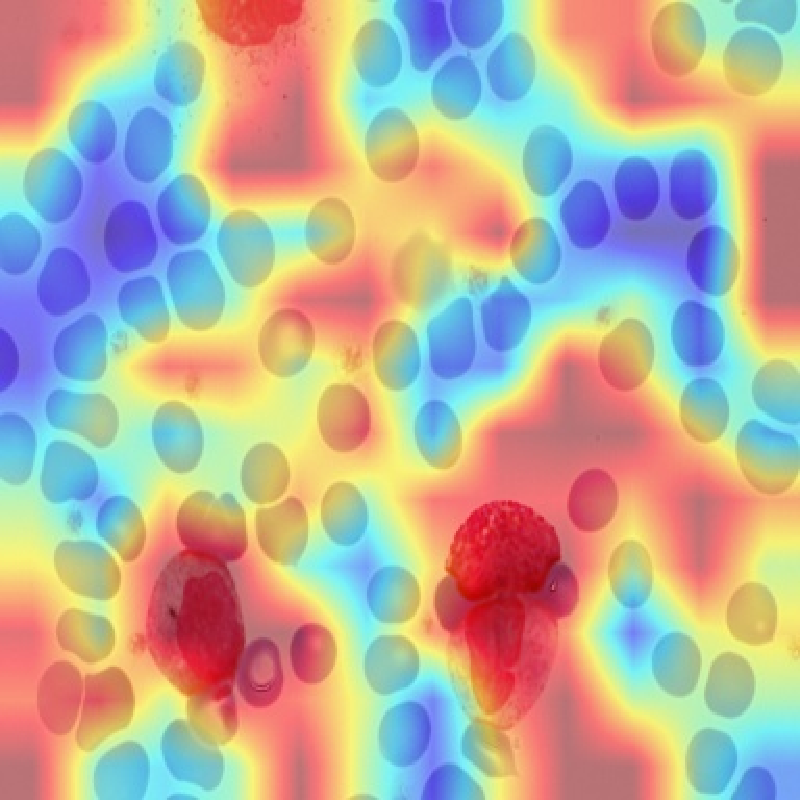}   &   
 \includegraphics[width=0.29\linewidth,height=2.5cm]{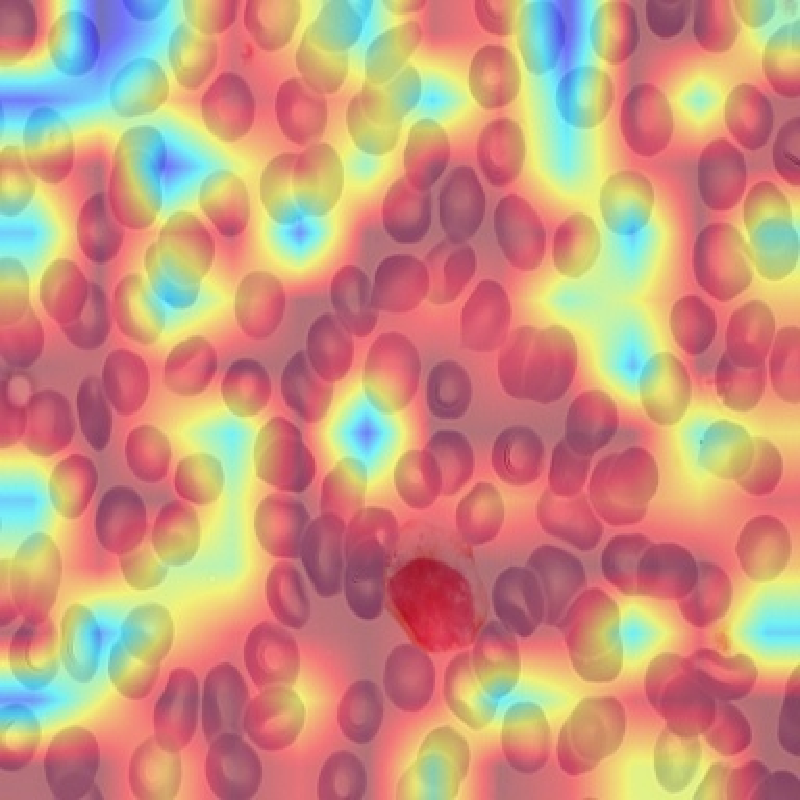} \\
\end{tabular}

\caption{Qualitative analysis of the ViT results using Grad-CAM. From left to right: CLL, CML, and Normal samples. From top to bottom: Input image, leaf classification output, and hierarchical classification output. The leaf model output has less discriminative distinction between WBCs and RBCs.}
\label{fig:grad_cam}
\end{figure}

To verify performance, Grad-CAM was used to visualize the ViT's  localization of features by targeting the final stage of the model and generating a heatmap. In Figure \ref{fig:grad_cam}, three sample input images (top row) are selected for testing with Grad-CAM.

It is expected that the model should perform well in identifying white blood cells (WBCs) from the images. Figure \ref{fig:grad_cam} shows the leaf classification model (middle row) identifying WBCs, but does not distinctively pick-out WBCs from the red blood cells (RBCs). However, the hierarchical model (bottom row) excels at pinpointing WBCs in every sample, signifying its higher effectiveness over the flat model. Yet, the hierarchical model displays certain limitations when interpreting Grad-CAM results. The heatmap spots appear wider in initial levels, becoming most precise at the lowest level of the hierarchy.

The authors believe this might be due to the model's multi-level hierarchical training, which necessitates a broader image area for decision-making. Additionally, it is observed that instances of lower model performance match with poorer localization in the heatmap. This is possibly due to visual similarities between classes, leading to the model's underperformance in these instances.

\section{Conclusion}

Our work demonstrates that multi-label hierarchical classification using ConvNext-Tiny and ViT-Small, holds great promise for leukemia subtype classification. Qualitative results further authenticated the model's reliability by showcasing its localization capabilities.

Naturally, this work is not without limitations. Early-stage leukemia may not be detectable via blood smear images. Other limitations to consider include the assumption of non-overlapping subgroups in the data.

For future research, we recommend gathering more data to enhance model performance. Moreover, studying inter and intra-observer variability could offer comparative insights against real pathologists. Furthermore, implementing newer algorithms and pre-training models on histology data are additional measures to consider for refining the classification process.

\textbf{Prospect of Application}: Leukemia is a curable disease; however, only if diagnosed properly. This work can be deployed to hospitals to reduce pathologists' errors, bias, and fatigue. It can also speed up the detection of leukemia subtypes and potentially save the lives of patients and the costs associated with cancer detection through a non-invasive technique.

\bibliography{mybibliography}

\end{document}